\documentclass[10pt,twocolumn,letterpaper]{article}

\usepackage{cvpr}              
\usepackage[accsupp]{axessibility}

\usepackage[dvipsnames]{xcolor}

\definecolor{cvprblue}{rgb}{0.21,0.49,0.74}
\usepackage[pagebackref,breaklinks,colorlinks,citecolor=cvprblue]{hyperref}
\usepackage{multirow}
\usepackage[normalem]{ulem}
\useunder{\uline}{\ul}{}

\def\paperID{1563} 
\def\confName{CVPR}
\def\confYear{2024}

\title{Separating the ``Chirp'' from the ``Chat'': \\ Self-supervised Visual Grounding of Sound and Language}

\author{Mark Hamilton\\
MIT, Microsoft\\
{\tt\small markth@mit.edu}
\and
Andrew Zisserman\\
Oxford, Google\\
\and
John R. Hershey\\
Google\\
\and
William T. Freeman\\
MIT, Google\\
}

\begin{document}
\twocolumn[{%
\renewcommand\twocolumn[1][]{#1}%
\maketitle

\begin{center}
    \centering
    \vspace{-.2in}
    \captionsetup{type=figure}
    \includegraphics[width=\textwidth]{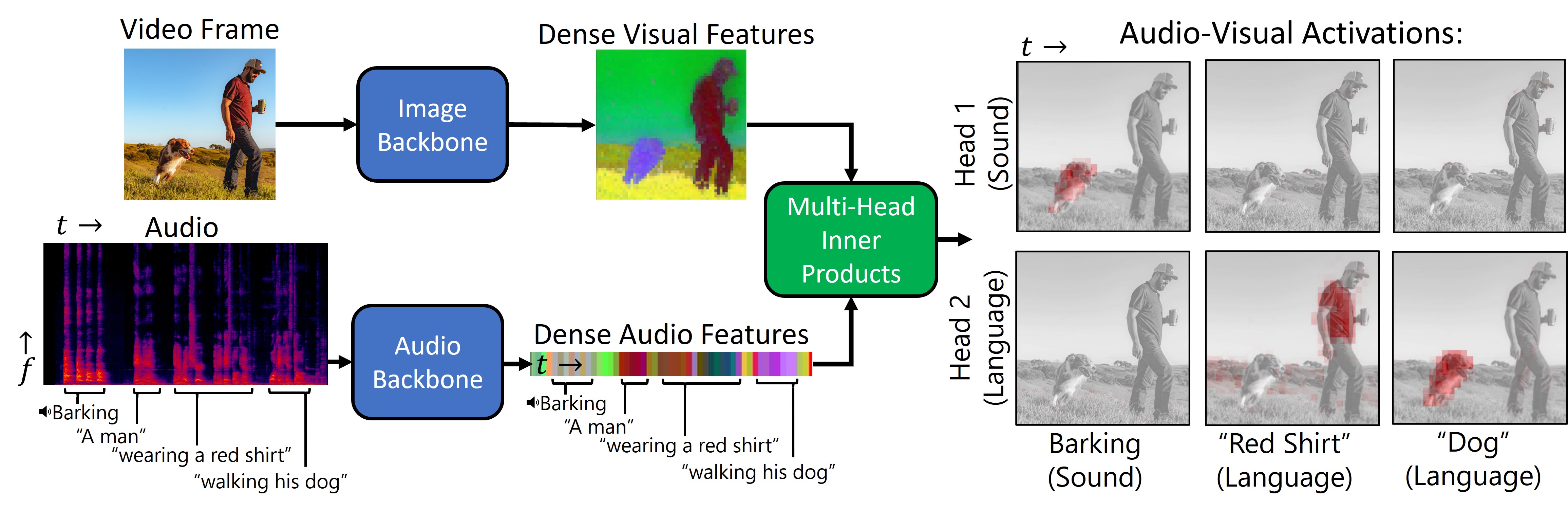}
    \vspace{-.3in}
    \captionof{figure}{Visual overview of the DenseAV algorithm. Two modality-specific backbones featurize audio and visual signals. We introduce a novel generalization of multi-head attention to extract attention maps that discover and separate the ``meaning'' of spoken words and the sounds an object makes. DenseAV performs this localization and decomposition solely through observing paired stimuli such as videos.
    }
    \label{fig:architecture}
\end{center}%
}]

\begin{abstract}

We present DenseAV, a novel dual encoder grounding architecture that learns high-resolution, semantically meaningful, and audio-visually aligned features solely through watching videos. We show that DenseAV can discover the ``meaning'' of words and the ``location'' of sounds without explicit localization supervision. Furthermore, it automatically discovers and distinguishes between these two types of associations without supervision. We show that DenseAV's localization abilities arise from a new multi-head feature aggregation operator that directly compares dense image and audio representations for contrastive learning. In contrast, many other systems that learn ``global'' audio and video representations cannot localize words and sound. Finally, we contribute two new datasets to improve the evaluation of AV representations through speech and sound prompted semantic segmentation. On these and other datasets we show DenseAV dramatically outperforms the prior art on speech and sound prompted semantic segmentation. DenseAV outperforms the previous state-of-the-art, ImageBind, on cross-modal retrieval using fewer than half of the parameters. Project Page: \href{https://aka.ms/denseav}{https://aka.ms/denseav}

\end{abstract}
    
\section{Introduction}
\label{sec:intro}

\begin{figure*}[t]
    \centering
    \vspace{-.25in}
    \includegraphics[width=\linewidth]{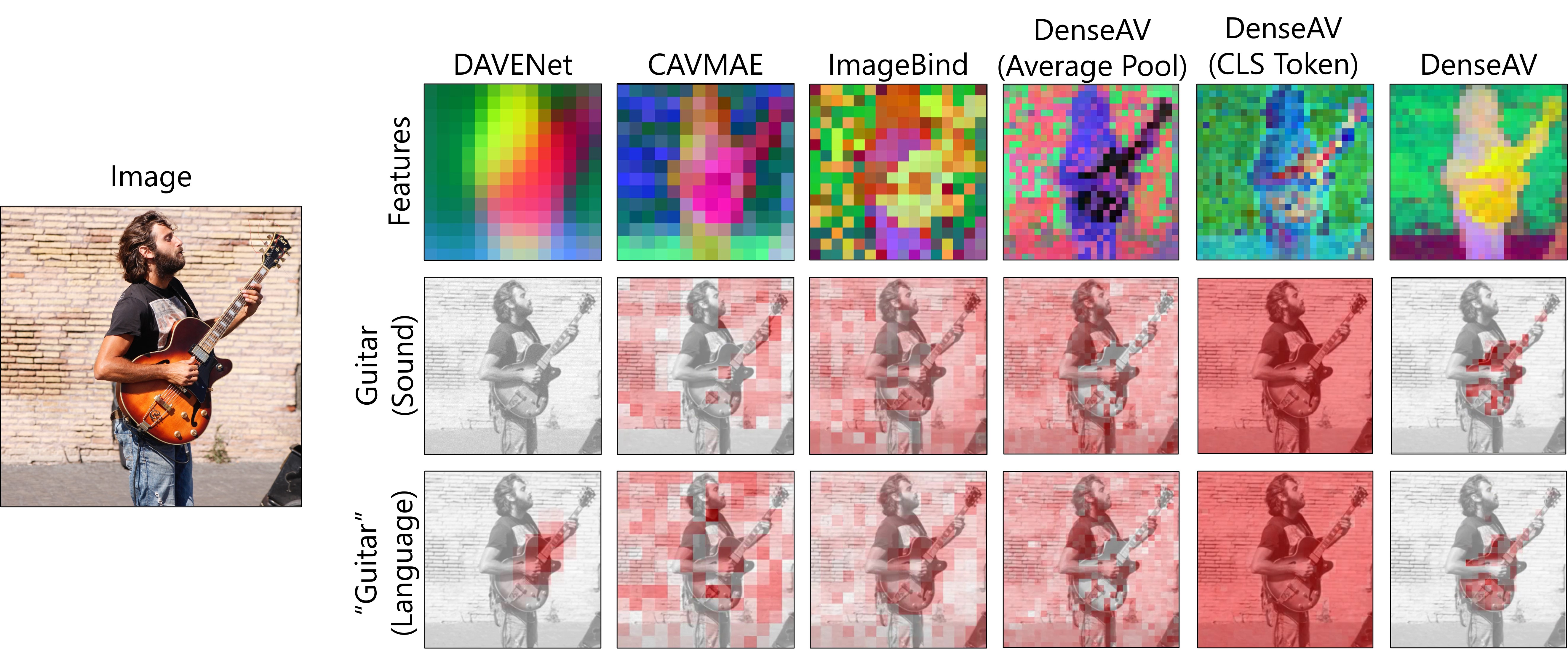}
    \vspace{-.25in}
    \caption{Qualitative comparison of several modern architectures for associating audio and video modalities. Only DenseAV learns a high-resolution and semantically aligned set of local features. This allows us to perform speech and sound prompted semantic segmentation using only the inner products between deep features. Other approaches, such as ImageBind, do not show aligned local feature maps. Approaches that do show some localization capabilities, like DAVENet, do not generalize to sound and language, and do not achieve the high-resolution localization capabilities of DenseAV. Dense features are visualized using PCA as in~\cite{hamilton2022unsupervised}}
    \label{fig:activation_comp}
    \vspace{-.1in}

\end{figure*}

Associating audio and video events is a fundamental task in human perception. As infants develop, the synchronization and correspondence of visible sounds enables multi-modal association -- a voice with a face, and a ``moo" with a cow~\cite{smith2008infants}. Later, as they acquire language, they associate spoken words with objects they represent~\cite{chomsky1987language,pullum2002empirical}. Amazingly, these association abilities, constituting speech recognition, sound event recognition, and visual object recognition, develop without much direct supervision. This work aims to create a model with this capability by learning high-resolution, semantically meaningful, audio-visually (AV) aligned representations. Features with these properties can be used to discover fine-grained correspondences between modalities without localization supervision or prior knowledge of the semantic representation of language. 

Consider the spoken caption and accompanying sounds of the image shown in Figure~\ref{fig:architecture}. We wish to ``ground'' both the speech and the sounds by identifying them with the corresponding visual objects. For instance, both the spoken word ``dog'' and the sound of a bark in the audio signal should be associated with the pixels of the dog in the visual signal if present. We seek high quality local representations where this behavior, which is notably absent from popular approaches in the literature,  emerges from simple inner products between cross-modal features.

To achieve this, we make three innovations. First, we introduce DenseAV, a dual-encoder architecture that computes a dense similarity volume over audio and visual features. 
Looking at a slice of this similarity volume for a spoken word, as in Figure~\ref{fig:architecture}, we can visualize the AV activation strength between a word or sound and an image's pixels. The novelty we introduce is to extend this dense similarity mechanism to have multiple similarity volume heads, much like those of multi-head attention. 
This allows each head to specialize on a particular type of coupling between the visual and audio modalities. Interestingly, we discover that if we give DenseAV two heads and train on a dataset that contains both language and sound, the heads naturally learn to distinguish language from more general sound using only cross-modal supervision. For example, as shown in Figure~\ref{fig:architecture}, head 1 focuses on sounds, such as a dog bark, emitted by visible objects, whereas head 2 focuses on speech, such as the word ``dog", that refers to visible objects.

Second, we show the importance of the ``aggregation function'' one uses to create a summary similarity score between an audio clip and a video frame for contrastive learning. The traditional choices, using inner products between global representations such as class tokens ~\cite{dosovitskiy2020image,caron2021emerging,shih2023speechclip} or pooled features~\cite{zhou2016learning,gong2022contrastive}, do not promote AV alignment of dense local features. Because of this, several popular audio-video backbones that excel on cross-modal retrieval \textit{cannot} directly associate objects and sounds using their local features. This limits their ability to be used for downstream tasks such as semantic segmentation, sound localization, or unsupervised language learning and discovery. 

Third, we introduce two semantic segmentation datasets to evaluate visual grounding with AV representations for speech and (non-speech) sounds. We build these datasets from the high-quality segmentation masks provided by the ADE20K dataset~\cite{zhou2017scene} and measure mean average precision (mAP) and mean intersection over union (mIoU) on a binary mask prediction task. This evaluation is simpler and more thorough than previous efforts to measure visual grounding such as the concept counting metrics of~\cite{harwath2018jointly} and the ``pointing games'' of~\cite{oquab2015object,arandjelovic2018objects,fong2017interpretable} that only check if a heatmap's peak occurs within a target box or segment. Furthermore, our evaluation avoids brittle word-net ontologies~\cite{miller1995wordnet}, clustering, Wu and Palmer distance~\cite{wu1994verb}, threshold choices, and a variety of other complicating factors. 

To summarize, our main contributions are as follows:

\begin{itemize}
    \item We introduce DenseAV, a novel self-supervised architecture that learns high-resolution AV correspondences.
    \item We introduce a local-feature-based image similarity function that significantly improves a network's  zero-shot localization ability  compared to common strategies such as average pooling or CLS tokens.
    \item We introduce new datasets for evaluating speech and sound prompted semantic segmentation. We show DenseAV significantly outperforms the current state-of-the-art on these tasks as well as on cross-modal retrieval.
    \item We discover that our multi-head architecture naturally disentangles audio-visual correspondence into sound and language components using only contrastive supervision.
\end{itemize}

\begin{figure*}[t]
    \centering
    \vspace{-.25in}
    \includegraphics[width=.9\linewidth]{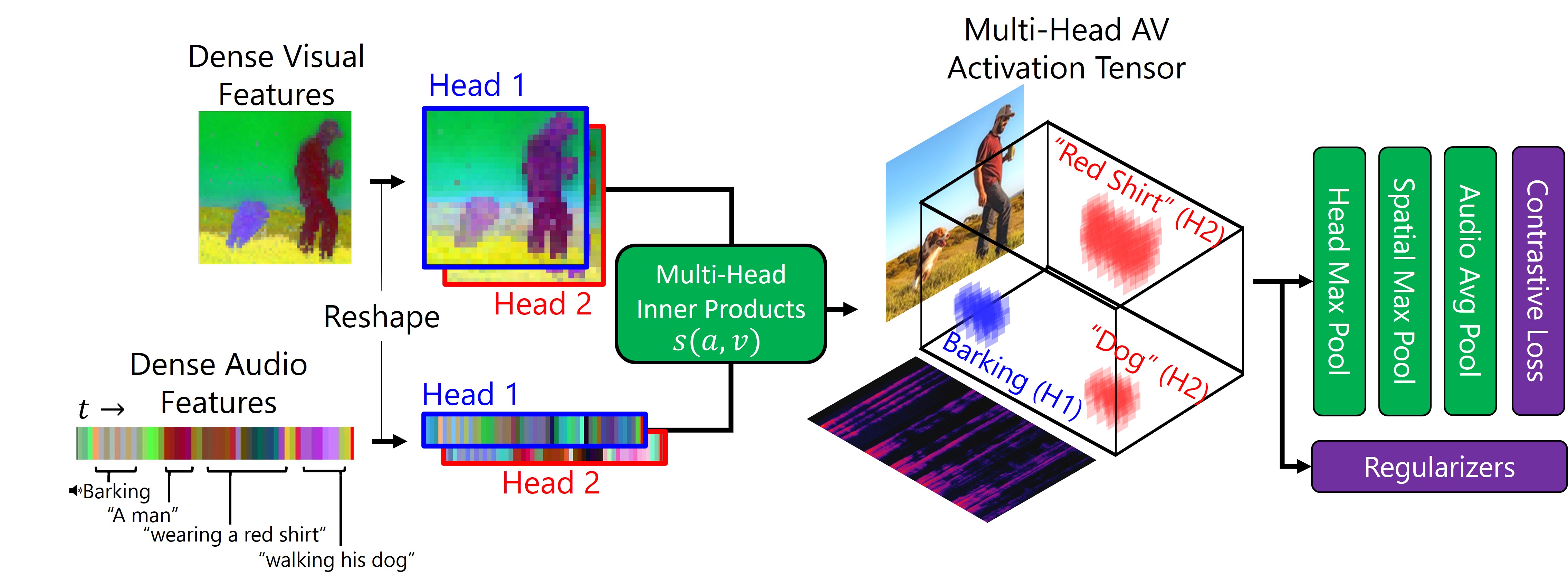}
    \vspace{-.12in}
    \caption{
    Architectural overview of our multi-head attention aggregator. Dense feature maps are split into $K$ heads $(K=1,2)$ in our experiments. We form an AV activation tensor by taking the inner-products of each head's features across the spatial and temporal extent of the visual and audio signals respectively as in Equation~\ref{eqn:pairedsim}. We then aggregate this similarity volume into a single similarity score by max-pooling head and spatial dimensions and average-pooling audio dimensions. Our approach aims to encourage the network to identify specific shared objects between the audio and visual modalities. In particular, max-pooling of heads disentangles sound and language, and max-pooling spatial dimensions helps localize objects. }
    \label{fig:loss_details}
    \vspace{-.1in}
\end{figure*}

\section{Related Work}
\label{sec:related}

Audio-visual (AV), text-visual, and other multi-modal models have a long history ~\cite{fisher2000learning,fisher2002probabalistic}, and have recently surged in popularity ~\cite{zhu2021deep}. Broadly speaking DenseAV is an audio-video contrastive learning architecture; this class of methods learns AV representations by aligning paired signals and pushing apart unpaired signals~\cite{chopra2005learning,jaiswal2020survey}. Of the models in this class, several stand out for their ability to localize sounds~\cite{arandjelovic2018objects,chen2021localizing,peng2022self} or capture the semantics of language~\cite{harwath2018jointly,peng2022word}. Many models in this class compare AV signals using inner products between ``global'' representations formed by pooled deep features~\cite{gong2022contrastive,wang2021multi,monfort2021spoken}, or class tokens~\cite{shih2023speechclip,peng2022self,peng2022word,girdhar2023imagebind,ma2020active}. Most notably, ImageBind has gained popularity due to its state-of-the-art performance on a variety of tasks and datasets and unified class-token-based contrastive architecture. In this work we show that many of these architectures do not show strong localization properties in their local features, despite excelling at cross-modal retrieval on a ``global'' level. This limits their applicability to new out-of-domain sounds, sounds that don't have a textual representation, and low-resource languages. We diverge from these works by directly supervising local tokens. In particular, we build on previous works~\cite{harwath2018jointly,arandjelovic2018objects} that show max-pooling improves localization capabilities and introduce a new multi-head aggregation operator that generalizes previous losses using a self-attention-like operator~\cite{vaswani2017attention}.  

Another class of methods discover structure in signals through uni- and multi-modal clustering. Early works on audio clustering~\cite{park2007unsupervised} discovered meaningful utterances without supervision. Similar visual analyses have discovered visual objects~\cite{ji2019invariant,cho2021picie,caron2018deep,hamilton2022unsupervised}. Recent works have applied these ideas to the AV domain ~\cite{harwath2017learning,alwassel2020self}, but do not focus on extracting \textit{high-resolution} AV representations. 

Finally, several works investigate generative audio-video learning. The Sound of Pixels~\cite{zhao2018sound} generates the sound of a specific object using a source separation loss. Newer approaches using GANs~\cite{kumar2020robust,liu2021towards}, and diffusion models~\cite{choi2023diffv2s,girdhar2023imagebind,mao2023contrastive} have generated audio from video and vice versa. Here we focus on improving the local representations of contrastive learners because of their relative scalability, simplicity, and ability to learn high-quality representations.

\section{Methods}
\label{sec:methods}

At a high level, DenseAV tries to determine when a given audio and visual signal belong ``together'' using dense audio-visual representations. To perform this task robustly, DenseAV must learn to predict the contents of an audio signal from a visual signal and vice versa. Doing so causes DenseAV to learn dense modality-specific features that capture the mutual information shared between the modalities \cite{tian2020contrastive}. Once learned, we can directly query these informative features to perform speech and sound prompted semantic segmentation as illustrated in Figure~\ref{fig:architecture}. 

More specifically, DenseAV is built from two modality-specific deep featurizers. These backbones produce temporally varying audio features across an audio clip and spatially varying video features for a single randomly selected frame. Our loss computes a similarity between audio and visual signals based on the intuition that two signals are similar if they have a variety of strong couplings or shared objects. More formally, we form a scalar similarity for a pair of audio and video signals by carefully aggregating a volume of pairwise inner products between dense features. We use the InfoNCE~\cite{oord2018representation} contrastive loss to encourage similarity between ``positive'' pairs of signals and dissimilarity between ``negative'' pairs formed by in-batch shuffling.  Figure~\ref{fig:loss_details} graphically depicts this loss function and subsequent sections detail each component of our architecture.

\subsection{Multi-Headed Aggregation of Similarities}

DenseAV's key architectural distinction is its loss function that directly supervises the ``local'' tokens of the visual and audio featurizers. This is a significant departure from other works~\cite{shih2023speechclip,guzhov2022audioclip,radford2021learning,caron2021emerging,oquab2023dinov2,girdhar2023imagebind} that pool modality specific information into ``global'' representations prior to the contrastive loss. Unlike prior works, our loss function aggregates the full pairwise similarities between the local tokens into an aggregate measure of similarity for a given pair of audio and visual signals. We show in Figure~\ref{fig:activation_comp} that this architectural choice enables DenseAV's local features to align across modalities whereas other approaches such as average pooling, class tokens, and SimPool \cite{psomas2023simpool} do not.

We first describe our loss function informally and definite it more precisely in the next paragraph. Our loss function computes the (un-normalized) inner product between every pair of visual and audio features to form a ``volume'' of inner products. This volume represents how strongly each part of an audio signal ``couples'' to each part of a visual signal. We aim to find many large couplings between positive pairs of audio and visual signals. Ideally, these couplings should connect visual objects with their references in the audio signal. Conversely, we do not want to find couplings between negative pairs of signals. To compute a single global coupling strength for a pair of signals, we aggregate this volume of pairwise similarities into a single number. There are myriad ways to aggregate this volume ranging from ``soft" average-pooling to ``hard" max-pooling. Average pooling yields dense gradients and can improve convergence speed and stability. However, max-pooling allows the network to focus on the \textit{best} couplings regardless the object's size or a sound's duration. Our aggregation function combines the benefits of average and max pooling by max-pooling visual dimensions and average pooling audio dimensions as proposed in~\cite{harwath2018jointly}. Intuitively speaking, this averages the strongest image couplings over an audio signal. It allows small visual objects to have large effects yet provides a strong training gradient to many regions of the signals. Finally, we draw inspiration from multi-head self-attention ~\cite{vaswani2017attention} and generalize this operation to multiple ``heads'' that we max-pool before pooling the visual and audio dimensions. This allows DenseAV to discover multiple ``ways'' to associate objects across modalities.

More formally, let $\mathcal{S}(a, v) \in \mathbb{R}$ represent the similarity between a tensor of audio features $a \in \mathbb{R}^{CKFT}$ of size (\textbf{C}hannel $\times$ \textbf{K}-heads $\times$ \textbf{F}requency $\times$ \textbf{T}ime) and a tensor of visual features $v\in \mathbb{R}^{CKHW}$ of size (\textbf{C}hannel $\times$ \textbf{K}-heads $\times$ \textbf{H}eight $\times$ \textbf{W}idth). To define this scalar similarity score, we first create a local similarity volume, $s(a,v) \in \mathbb{R}^{kfthw}$. For simplicity, we consider the aggregated similarity between a single image and audio clip but note one can easily generalize this to max-pool over video-frames. We define the full pairwise volume of similarities as:

\vspace{-.18in}
\begin{equation}
    s(a,v) \in \mathbb{R}^{kfthw} = \sum_{c=1}^{C} a[c,k,f,t] \cdot v[c,k,h,w]
    \label{eqn:pairedsim}
\end{equation}

Where $a[c,k,f,t]$ represents the value of $a$ at location $[c,k,f,t]$ and $\cdot$ is scalar multiplication. We aggregate this similarity volume into a single score $\mathcal{S}(a, v) \in \mathbb{R}$:

\vspace{-.18in}
\begin{equation}
    \label{eqn:sim_agg}
    \mathcal{S}(a, v) = \frac{1}{FT} \sum_{f=1}^{F} \sum_{t=1}^{T} \max_{k,h,w} \left( s(a,v)[k,f,t,h,w] \right) 
\end{equation}

 We note that this operation can be viewed as a multi-head generalization of the MISA loss of~\cite{harwath2018jointly}, and a multi-head multi-time generalization of the MIL loss of~\cite{arandjelovic2018objects}.

\subsection{Loss}

We can use the similarity between audio and visual signals defined in Equation~\ref{eqn:sim_agg} to construct a contrastive loss. We follow recent works~\cite{girdhar2023imagebind,frosst2019analyzing,wang2020understanding} and use the temperature-weighted InfoNCE ~\cite{oord2018representation} to encourage similarity between positive pairs of signals and dissimilarity between negative pairs. In DenseAV, we form $B$ positive pairs by splitting the audio and visual components of a $\textbf{B}$atch of training data. We form $B^2-B$ negative pairs by comparing a signal to all of the other signals in the training batch. More formally let $(a_b, v_b)_1^B$ be $B$ pairs of audio and visual signals. The visual-retrieval term of our InfoNCE loss is then:

\vspace{-.18in}
\begin{equation}
    \label{eqn:contrastive}
    \mathcal{L}_{A \to V} = \frac{1}{2B} \sum_{b=1}^{B} \left( \log \frac{ \exp{ \left( \gamma \mathcal{S}(a_b, v_b)  \right) }}{\sum_{b'=1}^{B} \exp{\left( \gamma\mathcal{S}(a_b, v_{b'})  \right) }} \right)
\end{equation}

Where $\gamma \in \mathbb{R}^+$ is a trainable inverse temperature parameter. We symmetrize this loss by adding the analogous audio-retrieval term, $\mathcal{L}_{V \to A}$, which iterates over negative \textit{audio} signals in the denominator.

\subsection{Audio and Visual Featurizers}

The core of DenseAV is two modality-specific backbone networks. We use the DINO vision transformer~\cite{caron2021emerging} with ImageNet pretrained weights (without labels) to provide a strong, yet fully unsupervised, vision backbone. Unlike other approaches that use CLIP~\cite{radford2021learning} as a backbone, DINO does not require paired text captions and learns from unlabeled images only. Practically, we find that DINO outperforms CLIP because of its better-behaved local tokens~\cite{darcet2023vision}, an effect we explore in the Supplement. We append an additional layer norm operation across the channel dimension~\cite{ba2016layer} and a $1\times1$ Convolution to DINO. The layer-norm and $1\times1$ convolution ensure the architecture does not start with a saturated loss function. We use the HuBERT audio transformer~\cite{hsu2021hubert} as DenseAV's audio backbone. HuBERT operates on waveforms and is trained on the LibriSpeech~\cite{panayotov2015librispeech} dataset using only self-supervision. Hubert outputs a single feature per time frame, corresponding to $F=1$ in Section~\ref{sec:methods}. Though HuBERT was only trained on speech, its audio features can be fine-tuned for more general sounds, much like how vision backbones can be fine-tuned for new datasets~\cite{yosinski2014transferable}. As in the visual branch, we append a channel-wise LayerNorm block and two $3\times3$ convolutions to the audio branch. These layers help the network avoid saturation and speed convergence. Furthermore, the two convolutions help the model aggregate information, which reduces the cost of the pairwise feature comparison used in our loss function. We refer to these added layers after the pretrained backbones as the ``aligners'' in later sections.

\subsection{Regularizers}

\noindent \textbf{Disentanglement Regularizer, $\mathcal{L}_{Dis}$}: We add a small regularization term to encourage each head of Equation~\ref{eqn:pairedsim} to specialize and learn independent types of audio-visual associations. Interestingly we find that our 2-head model naturally learns to distinguish the meaning of words with one head and capture the sounds objects produce with another head. To further encourage this unsupervised discovery of concepts, we penalize the network when multiple attention heads are simultaneously active. More precisely, let $(a_b, v_b)_1^B$ be a \textbf{B}atch of $B$ paired audio and visual signals. Our disentanglement loss for two heads is then:

\vspace{-.1in}
\begin{equation}
	\mathcal{L}_{Dis} =\text{Mean}( |s(a_b, v_b)[1] \circ s(a_b, v_b)[2]|)
\end{equation} 

Where $\circ$ is elementwise multiplication and $| \cdot |$ is the elementwise absolute value function. $[k]$ mirrors PyTorch slicing notation and refers to selecting the activations for only the $k$th attention head. Intuitively, this loss encourages one head to be silent if the other head is active and is a ``cross-term'' generalization of the $l^2$ regularizer~\cite{hoerl1970ridge} for encouraging activation shrinkage. When $K>2$ we average contributions from every combination of heads. We ablate this, and our decision to max-pool heads in Table~\ref{table:disentangle}.

\noindent \textbf{Stability Regularizers, $\mathcal{L}_{Stability}$}: Finally, we add several other small regularization terms to encourage stable convergence. We detail and ablate these terms in the Supplement. Briefly, these terms include standard regularizers like Total Variation~\cite{rudin1992nonlinear} smoothness over time and non-negative pressure to encourage the network to focus on similarity instead of dissimilarity. In addition, we add a regularizer to prevent the calibration temperature, $\gamma$, from drifting too quickly, and a regularizer to discourage activations during silence and noise. In the supplement we show that each regularizer alone does not have a dramatic effect on final metrics but together they can stop collapses during training.

Combining these losses into a single loss function yields:

\vspace{-.2in}
\begin{equation}
        \mathcal{L} = \mathcal{L}_{A \to V} + \mathcal{L}_{V \to A} + \lambda_{Dis}\mathcal{L}_{Dis} + \mathcal{L}_{Stability} 
\end{equation}

In our experiments we use $ \lambda_{Dis} = 0.05$ and refer interested readers to the supplement for the details of our small stability regularizer, $\mathcal{L}_{Stability}$.

\subsection{Training}

In our experiments we train DenseAV and relevant baselines on the AudioSet~\cite{gemmeke2017audio} dataset for sound prompted segmentation and AudioSet retrieval. We train on the PlacesAudio~\cite{harwath2016unsupervised} dataset for speech prompted segmentation, PlacesAudio retrieval, and the ablation studies of Table~\ref{table:ablation}. In our disentanglement experiments of Table~\ref{table:disentangle} and feature visualizations of Figures~\ref{fig:architecture} and ~\ref{fig:activation_comp} we train on both AudioSet and PlacesAudio so that DenseAV can be familiar with both language, the prominent audio signal in PlacesAudio, and more general sounds from AudioSet. In these experiments we sample training data from these two corpora, so each batch has an even split between AudioSet and PlacesAudio.

\noindent \textbf{Warming up Aligners:} We find that we can dramatically improve the stability by first training the added aligners (convolutions and layer norms) for $3000$ steps while keeping pretrained DINO and HuBERT backbones fixed. This allows the aligners to adapt to these intelligent backbones before modifying each backbone's sensitive weights. We use random resize crops, color jitter, random flips, and random greyscaling as image augmentations. We randomly sample a single video frame to feed to our visual branch. Audio clips are converted to single-channel format and are trimmed or padded with silence to create uniform 10 second clips. We re-sample audio clips according to the requirements of the backbone models used. For HuBERT, we re-sample to 16KhZ. We train on 8 V100 GPUs with an effective batch size of 80, and aggregate negative samples on all GPUs prior to computing the loss to ensure efficient parallelization. We provide additional training information and hyperparameters in the supplement.

\noindent  \textbf{Full Training:} After warming up the aligners, we train the full model for an additional 800,000 steps using the same loss, batch-size, and training logic. We train all aligner weights and fine-tune all HuBERT audio backbone weights. We use low rank adaptation (LoRA)~\cite{hu2021lora} to fine-tune the ``Q'', ``K'', and ``V'' layers of the DINO visual backbone attention blocks. This allows us to efficiently adapt DINO and stabilize the training as it is quite easy to collapse the carefully trained DINO weights. We use a LoRA rank of 8.

\begin{table}[t]
\vspace{-.25in}
\centering
\begin{tabular}{c|cc|cc}
\toprule
\multirow{2}{*}{Method} & \multicolumn{2}{c|}{Speech Semseg.} & \multicolumn{2}{c}{Sound Semseg.}  \\
                        & mAP              & mIoU            & mAP             & mIoU            \\ \midrule
DAVENet \cite{harwath2018jointly}               & {\ul 32.2\%}     & {\ul 26.3\%}    & 16.8\%          & 17.0\%          \\
CAVMAE \cite{gong2022contrastive}                 & 27.2\%           & 19.9\%          & {\ul 26.0\%}    & {\ul 20.5\%}    \\
ImageBind \cite{girdhar2023imagebind}              & 20.2\%           & 19.7\%          & 18.3\%          & 18.1\%          \\
\textbf{Ours}           & \textbf{48.7\%}  & \textbf{36.8\%} & \textbf{32.7\%} & \textbf{24.2\%} \\ \bottomrule
\end{tabular}
\vspace{-.1in}
\caption{
\textbf {Speech and Sound prompted semantic segmentation}. We analyze the quality of local features using two prompted semantic segmentation tasks. We prompt networks with speech of the form ``a picture of a(n) [Object]'' to determine whether local feature inner products can segment objects in the ADE20K dataset by name. We create sound prompts for a given ADE20K class using a curated mapping from the ADE20K ontology to the VGGSound ontology. DenseAV's local features perform significantly better than all baselines investigated. We bold ``first place'' results and underline ``second place'' results. 
}
\vspace{-.1in}

\label{tab:semseg}
\end{table}

\section{Experiments}
\label{sec:experiments}

To evaluate AV representation quality, we perform a variety of analyses including comparative activation visualization, quantitative measurements of speech and sound prompted semantic segmentation, and cross-modal retrieval. Additionally, we quantify our observation that DenseAV can distinguish the meanings of words (language), from the sounds of objects (sound) without supervision.

To adequately measure a representation's AV alignment quality, we found it necessary to introduce two evaluation datasets that measure speech and sound prompted semantic segmentation performance. Our two datasets introduce pairs of speech and sound prompts coupled with matching images and segmentation masks derived from ADE20K. We create these datasets because previous works~\cite{harwath2018jointly} have not published their datasets or evaluation code. However, we use an experimental setting from the literature for our cross-modal retrieval experiments.  

We compare against a variety of prior art including the popular state-of-the art multi-modal retrieval network, ImageBind~\cite{girdhar2023imagebind}. We also compare against CAVMAE~\cite{gong2022contrastive}, a leading multimodal backbone trained specifically for AudioSet retrieval, and DAVENet~\cite{harwath2018jointly}, which is trained to localize the meanings of words. We include two other baselines~\cite{harwath2016unsupervised,harwath2017learning} which have reported cross modal retrieval metrics on Places Audio. Finally, we compare our multi-head aggregation strategy to common ``global'' retrieval methods such as inner products between class-tokens, average-pooled tokens, and SimPooled\cite{psomas2023simpool} tokens. We note that SimPool achieves state-of-the-art localization results when compared to 14 other pooling methods. Nevertheless, our multi-head aligner yields better localization results than any of these ``global'' methods.

\subsection{Qualitative Comparison of Feature Maps}

Our first experiment in Figure~\ref{fig:activation_comp} highlights the dramatic differences in quality between DenseAV's features and other approaches in the literature. DenseAV is the only backbone whose local tokens are semantically meaningful and show cross-modal alignment for speech and sound. Though both CAVMAE and ImageBind show high-quality retrieval performance, neither shows high quality aligned local tokens. As a result, DenseAV can associate and localize both sound and language significantly better than other backbones. DAVENet shows coarse correspondences between language and visual objects but cannot associate sound with visual objects and does not match DenseAV's high resolution maps. Furthermore, the right half of Figure~\ref{fig:architecture} demonstrates that DenseAV naturally discovers and separates word semantics from the sound of objects without labels to supervise this separation. In the supplement, we provide additional visualizations of all backbones considered across a wide range of words and sounds.

\begin{table}[t]
\vspace{-.25in}
\centering
\begin{tabular}{c|cccc}
\toprule
\multirow{2}{*}{Method} & \multicolumn{2}{c}{Places Acc. @10} & \multicolumn{2}{c}{AudioSet Acc. @10} \\
                        & I $\rightarrow$ A       & A $\rightarrow$ I      & I $\rightarrow$ A    & A $\rightarrow$ I   \\ \midrule
\cite{harwath2016unsupervised}*              & 46.3\%                  & 54.8\%                 & -                    & -                   \\
\cite{harwath2017learning}*              & 54.2\%                  & 56.4\%                 & -                    & -                   \\
DAVENet \cite{harwath2018jointly}*                 & 52.8\%                  & 60.4\%                 & -                    & -                   \\

CAVMAE \cite{gong2022contrastive}                & {\ul 81.7\%}            & {\ul 77.7\%}           & 55.7\%               & 50.7\%              \\
ImageBind \cite{girdhar2023imagebind}              & 1.10\%                   & 1.10\%                  & {\ul 64.5\%}         & {\ul 66.5\%}        \\
\textbf{Ours}           & \textbf{94.2\%}         & \textbf{94.3\%}        & \textbf{69.8\%}      & \textbf{68.1\%}     \\ \bottomrule
\end{tabular}
\vspace{-.1in}
\caption{
\textbf{Cross-modal retrieval using 1000 evaluation videos from the PlacesAudio and AudioSet validation datasets}. DenseAV dramatically outperforms all approaches tested in all metrics. Most notably, the state-of-the-art image retrieval foundation model, ImageBind, is incapable of recognizing speech. We note that the ImageBind authors do not publish retraining code, so we evaluate their largest pretrained model. Models with a * indicate that they have been previously reported in the literature. Other numbers are calculated by using pretrained models when available or from training with the author's official training scripts.
}
\label{tab:retrieval}
\end{table}

\subsection{Speech Prompted Image Segmentation}

\label{sec:speech-prompted}

\textbf{Dataset:} We introduce a speech prompted segmentation dataset using the ADE20K dataset, which is known for its comprehensive ontology and pixel-precise annotations~\cite{zhou2017scene}. From this dataset, we curate an evaluation subset of image-class pairs by sampling up to 10 images for each object class in ADE20K, excluding images where the selected class was tiny ($< 5\%$ of pixels). We only consider classes with at least 2 images that pass the tiny object criterion. For each class and image, we formed a binary target mask by selecting the semantic segmentation mask for that class. This resulted in 3030 image-object pairs spanning 478 ADE20K classes.

We created paired speech signals by speaking the prompt ``A picture of a(n) [object]'' where [object] is the name of the ADE20K class. We create clear, controlled, and consistent audio prompts using Microsoft's neural text to speech service~\cite{ren2020fastspeech}. This service also provides exact timing of the ``[object]'' utterance within the broader prompt and ensures each class is measured equally. Grammar was manually verified for the utterances to ensure proper singular/plural and a/an agreement with the class name. We release images, masks, and audio prompts for reproducibility.

\noindent \textbf{Evaluation Measure:} We evaluate methods based on how well their speech-prompted activations align with ground truth masks for the visual object's class. We quantify this with the binary Average Precision (AP) and Intersection over Union (IoU) metrics. These quantify how close activations match with the binary label mask from the ADE20K dataset. To compute an aggregate score over all of the object classes considered, we compute the mean average precision (mAP) and mean intersection over union (mIoU) by averaging AP scores across all object categories considered. 

The mAP is particularly well suited for evaluating feature similarities because it is unaffected by monotonic transformations of the similarity scores. This eliminates the need for arbitrary thresholding and calibration. This is particularly important because many networks’ inner products are not centered at zero, and the best thresholding strategy can be nontrivial, and dependent on the network and object class. Average Precision avoids these confounding factors and ensures a fair comparison across methods. Unfortunately, unlike the mAP, the mIoU metric requires selecting a threshold. To ensure our mIoU measurement is similarly invariant to monotonic transformations we evaluate 20 uniformly spaced thresholds between the smallest and largest activations of each model. For each baseline, we report results for the best threshold to ensure a fair comparison between all networks considered.

\noindent \textbf{Implementation:} We compute image heatmaps by evaluating each modality-specific network on the image-audio pairs from our dataset. We extract dense features from the final layer of each network and form their similarity volume according to Equation~\ref{eqn:pairedsim}. For DenseAV we max-pool the head dimension to properly compare with single-headed models. We average activations over the temporal extent of the ``[object]'' utterance using the word timing information from the ground truth audio clip. This creates a heatmap over the image features that can be bi-linearly resized to the original image's size. We then compare these per-pixel activation scores to ground truth object masks from our dataset.

\noindent \textbf{Results:} In Speech mAP and mIoU columns of Table~\ref{tab:semseg} we show that DenseAV achieves a \textbf{51\% (+16.5 mAP)} relative increase in speech-prompted semantic segmentation over previous methods. Approaches that use global token based contrastive strategies such as CAVMAE and ImageBind perform particularly poorly in this task, and this observation aligns with the qualitative results of Figure~\ref{fig:activation_comp}.

\begin{table}[t]
\vspace{-.25in}
\centering
\begin{tabular}{c|cc}
\hline
Method                                & Pred. Dis.         & Act. Dis.       \\ \hline
No $\mathcal{L}_{Dis}$, No Head Max Pool             & 64.1\%          & 70.3\%          \\
No $\mathcal{L}_{Dis}$      & \textbf{99.9\%} & \underline{86.5}\%          \\
\textbf{Ours}                         & \textbf{99.9\%} & \textbf{91.2\%} \\ \hline
\end{tabular}
\vspace{-.1in}
\caption{Quantitative ablation study of the impact of max-pooling attention heads and adding our disentanglement loss, $\mathcal{L}_{Dis}$. Intuitively, max-pooling attention heads allows each head to specialize on its own specific set of triggers. Our disentanglement loss further encourages the heads to operate independently and orthogonally. }

\label{table:disentangle}
\end{table}

\subsection{Sound Prompted Image Segmentation}

\textbf{Dataset:} To evaluate how well deep features localize sound, we build on Section~\ref{sec:speech-prompted} and create a dataset of sound prompts that align with ADE20K classes. We first select the same (large) image-object pairs from ADE20K. We then create a mapping between the ADE20K and VGGSound~\cite{chen2020vggsound} ontologies. To compute a robust mapping, we first embed ADE20K class names and VGGSound class names with the GPT Ada 2 text embedding model~\cite{openai2023gpt4}. For each ADE20K class, we create a list of at most three candidates from the VGGSound ontology that have a cosine similarity $(>.85)$. We then manually review these candidates to select the best VGGSound class for each ADE20K class and remove any spurious or mistaken matches. This produces a set of 95 ADE20K classes with strong matches in the VGGSound ontology. For each of our original 3030 image-object pairs we select a random VGGSound validation clip with a matching class according to our mapped ontology. This yields 106 image-object pairs across 20 ADE20K classes. 

\noindent \textbf{Evaluation Measure:} We use the same mAP and mIoU evaluation metrics as Section~\ref{sec:speech-prompted}, but instead average over the 20 ADE20K classes considered.

\noindent \textbf{Implementation:} We compute sound prompted image activations as in section~\ref{sec:speech-prompted} but with one key change: we average activations over the entire clip because we do not have ground-truth sound timing information. 

\noindent \textbf{Results:}  The ``Sound mAP and mIoU'' columns of Table~\ref{tab:semseg} show that DenseAV achieves a $ \textbf{25\%}$ $ \textbf{(+6.4\text{mAP}) }$ relative improvement in sound prompted segmentation compared to the prior art. Most notably, ImageBind's features cannot localize sound despite their high cross-modal retrieval performance learned from millions of hours of sound.

\begin{table}[t]
\vspace{-.25in}
\centering
\begin{tabular}{c|ccc}
\hline
\multirow{2}{*}{Method}                     & \multirow{2}{*}{Speech mAP}      & \multicolumn{2}{c}{Places Acc. @10}                                 \\
                                            &                                  & V $\rightarrow$ A                & A $\rightarrow$ V                \\ \hline
Average Pool                                & 20.1\%                           & 92.0\%                           & 91.2\%                           \\
CLS Token                                   & 20.6\%                           & 86.4\%                           & 89.8\%                           \\
SimPool \cite{psomas2023simpool}                                   & \underline{35.3}\%                           & \underline{92.6}\%                           & \underline{92.8}\%                           \\
\textbf{Multi-Head (Ours)} & \textbf{48.2\%} & \textbf{93.5\%} & \textbf{93.8\%} \\ \hline
\end{tabular}
\vspace{-.1in}
\caption{Quantitative ablation of different feature aggregation strategies. Though the common practice of average pooling and using a learned CLS token to aggregate features have little effect on retrieval performance, they dramatically degrade performance on speech prompted semantic segmentation. }
\label{table:ablation}

\end{table}

\subsection{Cross-Modal Retrieval}

We show that DenseAV's representations are not only better for localization, but significantly outperform other approaches on cross-modal retrieval. We adopt the evaluation setting of~\cite{harwath2018jointly} and measure cross modal retrieval accuracy at 1, 5, and 10 in a thousand-way retrieval task. In particular, we use the same thousand images from the validation set of~\cite{harwath2018jointly} and also replicate this analysis on one-thousand random clips from the AudioSet validation data. Table~\ref{tab:retrieval} shows results for 1000-way retrieval tasks on both the Places Audio and AudioSet datasets. We show cross-modal accuracy at 10, but also show larger tables in the supplement that echo these results using accuracy at 1 and 5. DenseAV significantly outperforms all baselines across all metrics. Interestingly, DenseAV outperforms ImageBind with \textit{less than half} of the trainable parameters and no reliance on text.

\subsection{Measuring Disentanglement}

We observe that DenseAV's heads naturally learn to differentiate audio-visual couplings that capture the meaning of words (language) and those that capture the sounds of objects (sound). Furthermore this effect generalizes to novel clips, including those with both sound and language as shown in Figure~\ref{fig:architecture}. We quantify this observation in two ways, the first measures if a head's average activation strength predicts whether a clip contains mainly ``language'' or ``sound''. The second method quantifies how often the ``sound'' head is incorrectly active when the ``language'' head should be active and vice versa. We leverage the fact that AudioSet dataset contains mostly clips with ambient sound and rarely contains language. In contrast, Places Audio is entirely language-based without external ambient sound. We note that these analyses are specifically for our architecture with two heads $K=2$ and trained on both AudioSet and PlacesAudio data.

For both measures of disentanglement, we first compute a clip's aggregated similarity for each head. In particular, we remove the max-pooling over heads in Equation~\ref{eqn:sim_agg} to create a single-head similarity, $\mathcal{S}(a,v)_k$. We then min-max scale the scores of each head across both datasets to lie in the $[0,1]$ interval, which we refer to as $\hat{\mathcal{S}}(a,v)_k$. Using these normalized scores, we can create metrics that capture how well a given head responds only to a specific dataset.

Our first metric measures how well a head's scores predict whether a clip is from the ``sound'' or ``language'' dataset. Let $(a_b, v_b)_1^B$ be tuples of paired audio and visual signals. let $l[k']_b$ be an indicator variable of whether the signal $(a_b, v_b)$ arises from the sound dataset, AudioSet, $(k'=1)$, or the language dataset Places Audio $(k'=2)$.

\vspace{-.1in}
\begin{equation}
	\delta_{pred}(k, k') = \text{AP}\left( (\hat{\mathcal{S}}(a_b,v_b)_k)_1^B, (l[k']_b)_1^B \right)
\end{equation}

Where $AP(\cdot, \cdot)$ is the binary average precision with prediction and label arguments respectively. Intuitively, this measures whether the scores of head $k$ are direct predictors of whether the data is from dataset $k'$. We can find the best assignment between heads and datasets such that each head is maximally predictive of the given dataset:

\vspace{-.2in}
\begin{multline}
    \label{eqn:pred-dis}
	\text{PredDis} = \frac{1}{2} \max \left( \delta_{pred}(0, 0) + \delta_{pred}(1, 1), \right. \\
	\left. \delta_{pred}(1, 0) + \delta_{pred}(0, 1) \right)
\end{multline}
\vspace{-.15in}

The prediction disentanglement score, PredDis, is a percentage that ranges from $50\%$ for completely entangled signals to $100\%$ if one can perfectly classify the signals using the scores of either head. The maximum over the two possible assignments makes this metric invariant to permutations of the heads. We note that this metric is a Hungarian matching assignment~\cite{kuhn1955hungarian} over two entries, a common technique to asses unsupervised classification performance~\cite{ji2019invariant,hamilton2022unsupervised}.

Our second measure quantifies ``spurious activations'' in the non-dominant head. A truly disentangled system should have a head that only fires on sound, and another head that only fires on language. We create another disentanglement measure, ActDis, by replacing $\delta_{pred}$ in Equation~\ref{eqn:pred-dis} with:

\vspace{-.18in}
\begin{equation}
	\delta_{act}(k, k') = 1 - \frac{1}{\sum_{b'} l[k']_{b'}} \sum_{b=1}^{B} \hat{\mathcal{S}}(a_b,v_b)_k \cdot l[k']_b
\end{equation}

Intuitively, this measures the ``inactivity'' of head $k$ on dataset $k'$. If head $k$ is totally silent on dataset $k'$ then $\delta_{act}(k, k') = 1$. Like PredDis, ActDis is a percentage ranging from $50\%$ to $100\%$ with $100\%$ representing perfect disentanglement where the sound head is completely silent during the language clips, and vice versa.

Table~\ref{table:disentangle} shows that DenseAV achieves near perfect predictive ($99\%$) and activation ($91\%$) disentanglement. It also shows that our disentanglement regularizer and max-pooling over heads improves DenseAV's natural ability to distinguish sound from language without supervision.

\section{Conclusion}
\label{sec:conclusion}

We presented DenseAV, a novel contrastive learning architecture that can discover the meaning of words and localize the sounds of objects using only video supervision. We are the first to observe both qualitatively and quantitatively that it’s possible to disentangle the meaning of words from the sound of objects with only a contrastive learning signal. DenseAV's success stems from its novel multi-head attention aggregation mechanism that encourages its modality-specific backbones to create high-resolution, semantically meaningful, and AV aligned representations. These properties of DenseAV's representation are not seen in other state-of-the-art models in the literature. Consequently, DenseAV significantly surpasses other leading models in dense prediction tasks such as speech and sound-prompted semantic segmentation as well as in cross-modal retrieval.

\section*{Acknowledgements}
\vspace{-.1in}

We would like to thank the Microsoft Research Grand Central Resources team for their gracious help performing the experiments in this work. Special thanks to Oleg Losinets and Lifeng Li for their consistent, gracious, and timely help, debugging, and expertise. Without them, none of the experiments could have been run.

We would also like to thank David Harwath, Andrew Rouditchenko, Yuan Gong, Didac Suris, Adria Recasens Continente, and Jim Glass for their help in running DAVENet and CAVMAE baselines and evaluations as well as for many helpful tips on audio visual contrastive learning.

This material is based upon work supported by the National Science Foundation Graduate Research Fellowship under Grant No. 2021323067. Any opinion, findings, and conclusions or recommendations expressed in this material are those of the authors(s) and do not necessarily reflect the views of the National Science Foundation. 
This work is supported by the National Science Foundation under Cooperative Agreement PHY-2019786 (The NSF AI Institute for Artificial Intelligence and Fundamental Interactions, http://iaifi.org/). This work is funded by a Royal Society Research
Professorship RSRP$\backslash$R$\backslash$241003, and EPSRC Programme Grant VisualAI EP/T028572/1.

{
    \small
    \bibliographystyle{ieeenat_fullname}
    \bibliography{main}
}

\clearpage
\setcounter{page}{1}
\maketitlesupplementary


\section{Full Cross Modal Retrieval Results}

\setlength{\tabcolsep}{3.5pt}
\begin{table*}[h]
\centering
\begin{tabular}{c|cccccc|cccccc}
\toprule
           & \multicolumn{6}{c|}{Places Audio Retrieval}                                                                                    & \multicolumn{6}{c}{AudioSet Retrieval}                                                                                         \\ \cline{2-13} 
           & \multicolumn{3}{c|}{I $\rightarrow$ A}                                   & \multicolumn{3}{c|}{A $\rightarrow$ I}              & \multicolumn{3}{c|}{I $\rightarrow$ A}                                   & \multicolumn{3}{c}{A $\rightarrow$ I}               \\  
Method     & @1              & @5              & \multicolumn{1}{c|}{@10}             & @1              & @5              & @10             & @1              & @5              & \multicolumn{1}{c|}{@10}             & @1              & @5              & @10             \\ \midrule
\cite{harwath2016unsupervised} & 12.1\%          & 33.5\%          & \multicolumn{1}{c|}{46.3\%}          & 14.8\%          & 40.3\%          & 54.8\%          & -               & -               & \multicolumn{1}{c|}{-}               & -               & -               & -               \\
\cite{harwath2017learning} & 13.0\%          & 37.8\%          & \multicolumn{1}{c|}{54.2\%}          & 16.1\%          & 40.4\%          & 56.4\%          & -               & -               & \multicolumn{1}{c|}{-}               & -               & -               & -               \\
DAVENet~\cite{harwath2018jointly}   & 12.7\%          & 37.5\%          & \multicolumn{1}{c|}{52.8\%}          & 20.0\%          & 46.9\%          & 60.4\%          & -               & -               & \multicolumn{1}{c|}{-}               & -               & -               & -               \\
DAVENet*~\cite{harwath2018jointly}   & 13.3\%          & 38.3\%          & \multicolumn{1}{c|}{51.2\%}          & 20.5\%          & 45.3\%          & 57.2\%          & 0.10\%          & 0.70\%          & \multicolumn{1}{c|}{1.30\%}          & 0.10\%          & 0.30\%          & 1.20\%          \\
CAVMAE*\cite{gong2022contrastive}    & {\ul 36.7\%}    & {\ul 70.3\%}    & \multicolumn{1}{c|}{{\ul 81.7\%}}    & {\ul 33.9\%}    & {\ul 65.7\%}    & {\ul 77.7\%}    & 22.8\%          & 44.9\%          & \multicolumn{1}{c|}{55.7\%}          & 21.1\%          & 41.7\%          & 50.7\%          \\
ImageBind\cite{girdhar2023imagebind}  & 0.10\%          & 0.50\%          & \multicolumn{1}{c|}{1.10\%}          & 0.10\%          & 0.40\%          & 1.10\%          & {\ul 29.6\%}    & {\ul 55.4\%}    & \multicolumn{1}{c|}{{\ul 64.5\%}}    & {\ul 31.8\%}    & {\ul 57.3\%}    & {\ul 66.5\%}    \\
Ours       & \textbf{65.3\%} & \textbf{90.0\%} & \multicolumn{1}{c|}{\textbf{94.2\%}} & \textbf{64.4\%} & \textbf{89.4\%} & \textbf{94.3\%} & \textbf{35.1\%} & \textbf{58.0\%} & \multicolumn{1}{c|}{\textbf{68.2\%}} & \textbf{33.6\%} & \textbf{59.3\%} & \textbf{68.4\%} \\ \bottomrule
\end{tabular}

\caption{Full cross modal retrieval results using the same setting of Table~\ref{tab:retrieval}. We note DenseAV outperforms all baselines in all metrics and all datasets.}

\end{table*}

\section{VGGSound Source Evaluation} 

Table~\ref{tab:vggss} adds evaluations on the VGGSound Source dataset. We note that VGGSS annotation's large bounding boxes do not reward high-resolution results. Nevertheless, DenseAV outperforms all methods including 5 additional baselines (Attention10K \cite{senocak2018learning}, AVObject \cite{afouras2020selfsupervised}, LVS \cite{chen2021localizing}, FNAC AVL \cite{sun2023learning}, and SLAVC \cite{mo2022closer}).

\begin{table}[h]
\centering
\begin{tabular}{c|cc}
\toprule
Method        & cIoU            & AUC             \\ \midrule
DAVENet       & 6.8\%           & 21.2\%          \\
CAVMAE        & 7.9\%           & 25.0\%          \\
ImageBind     & 3.4\%           & 20.5\%          \\
Attention10K  & 18.5\%          & 30.2\%          \\
AVObject      & 29.7\%          & 35.7\%          \\
LVS           & 34.4\%          & 38.2\%          \\
SLAVC         & 38.8\%          & 38.8\%          \\
FNAC AVL      & {\ul 39.4\%}    & {\ul 39.4\%}    \\
\textbf{Ours} & \textbf{40.6\%} & \textbf{40.6\%} \\ \bottomrule
\end{tabular}
\caption{Performance on VGGSound Source localization.}
\label{tab:vggss}

\end{table}

\newpage

\section{Speech Prompted Semantic Segmentation Noise Robustness:} DenseAV was trained with natural speech and sounds and is robust to environmental noise and common speech errors like stutters. We explore additional noise-robustness experiments in Table \ref{tab:noise}. 

\begin{table}[h]
\centering
\begin{tabular}{c|cc}
\toprule
Method        & mAP             & mIoU            \\ \midrule
DAVENet      & {\ul 31.8\%}    & {\ul 26.1\%}    \\
CAVMAE       & 27.2\%          & 23.8\%          \\
ImageBind     & 20.2\%          & 19.7\%          \\
\textbf{Ours} & \textbf{48.1\%} & \textbf{36.6\%} \\ \bottomrule
\end{tabular}
\caption{Performance on speech based semantic segmentation task with environmental noise from the MUSAN \cite{snyder2015musan} dataset added to spoken category labels.}
\label{tab:noise}
\end{table}

\section{Speech Prompted Semantic Segmentation Examples}


\begin{figure}[h]
    \centering
    \includegraphics[width=\linewidth]{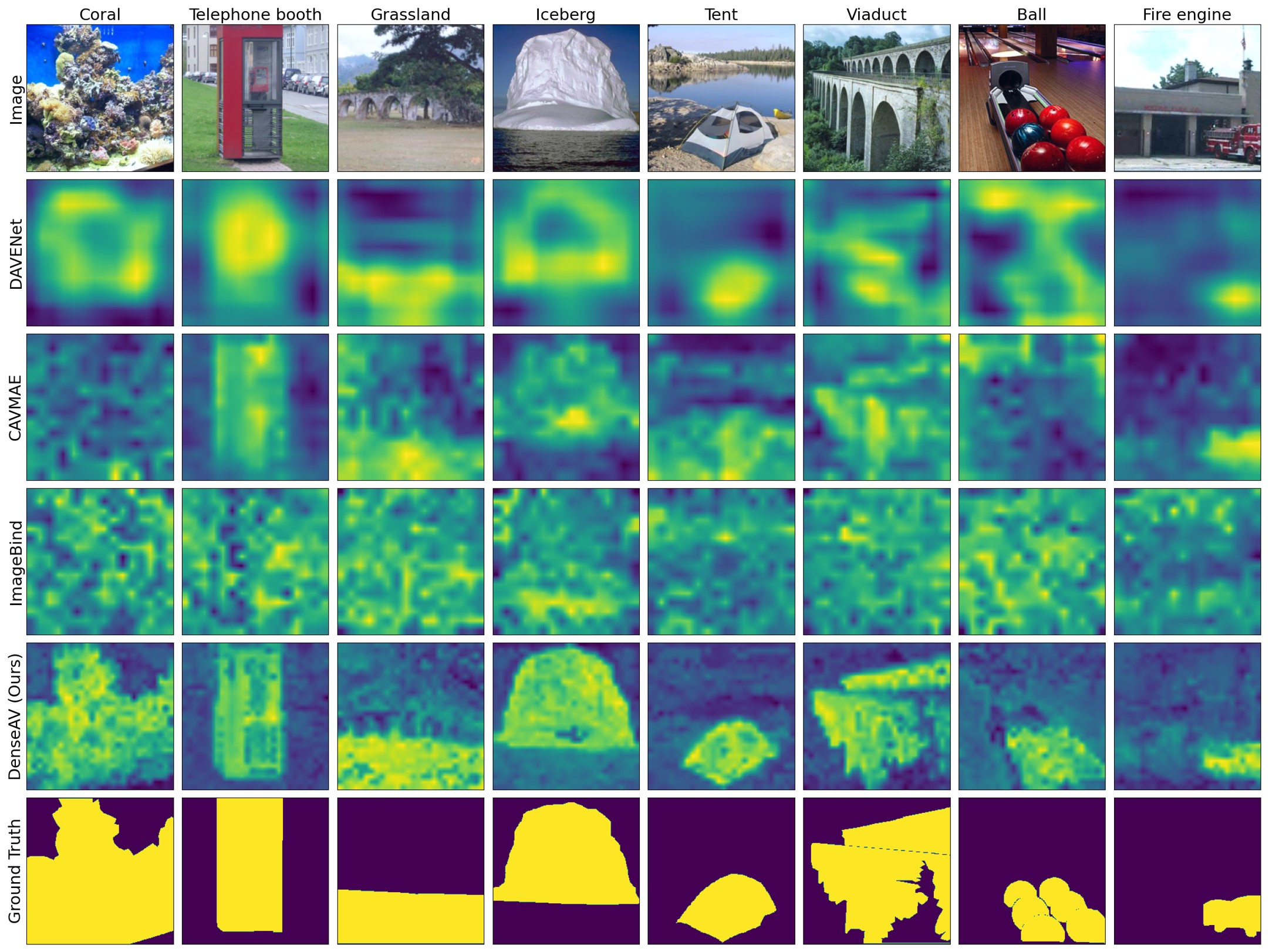}
    \caption{Selected visualizations of AV heatmaps for the {\em speech prompted} semantic segmentation task. We visualize results across several baselines. DenseAV achieves the best localization performance both qualitatively and quantitatively, highlighting the full extent of objects with high resolution heatmaps. }
    \label{fig:additional_speech}
\end{figure}

\newpage

\section{Sound Prompted Semantic Segmentation Examples}

\begin{figure}[h]
    \centering
    \includegraphics[width=\linewidth]{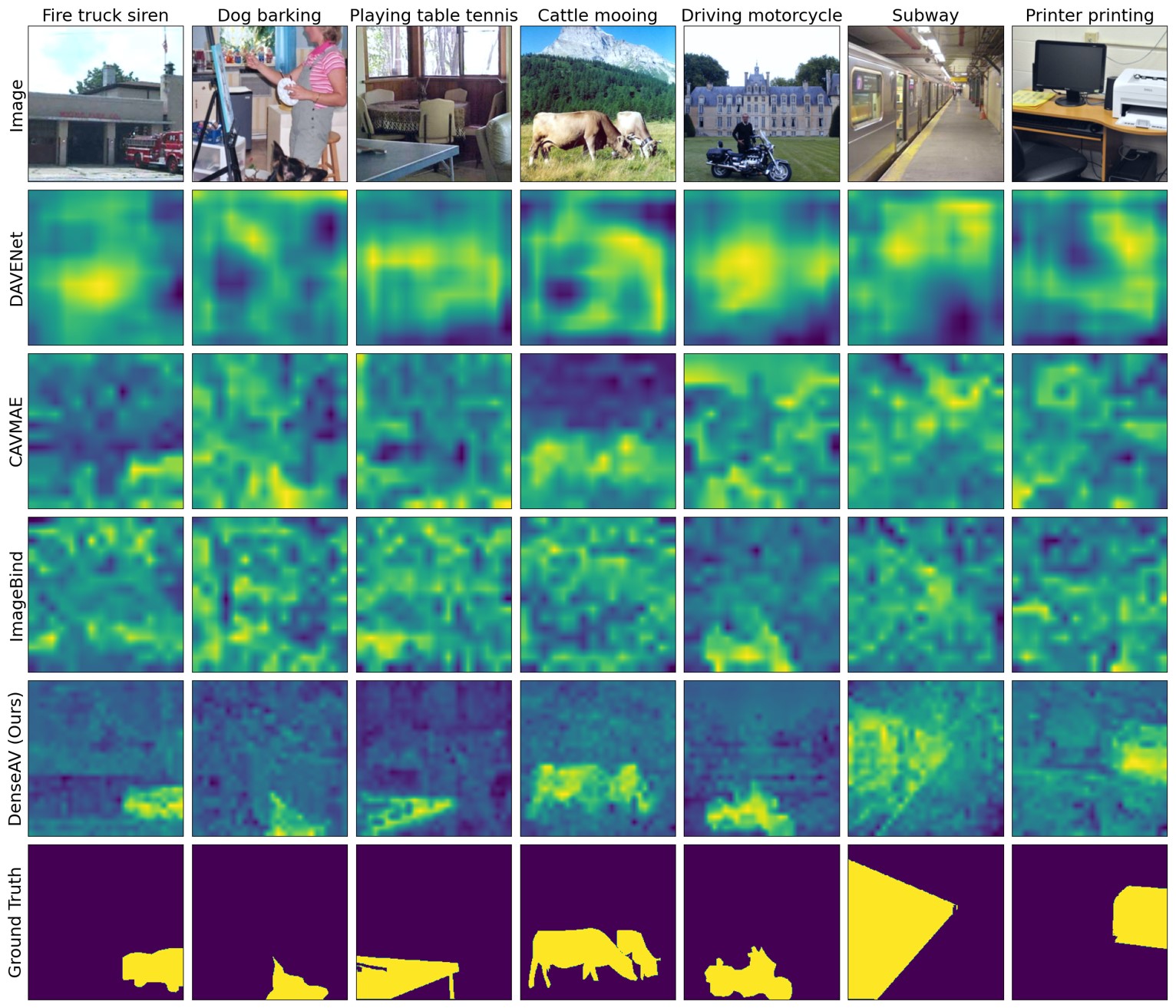}
    \caption{Selected visualizations of AV heatmaps for the {\em sound prompted} semantic segmentation task. We visualize results across several baselines. DenseAV achieves the best localization performance both qualitatively and quantitatively, highlighting the full extent of objects with high resolution heatmaps. We note that DenseAV can highlight objects even if they are not centered or clearly visible as in the dog example (second column).}
    \label{fig:additional_sound}
\end{figure}



\newpage

\section{Comparison Across Backbones}

\begin{figure}[h]
    \centering
    \includegraphics[width=\linewidth]{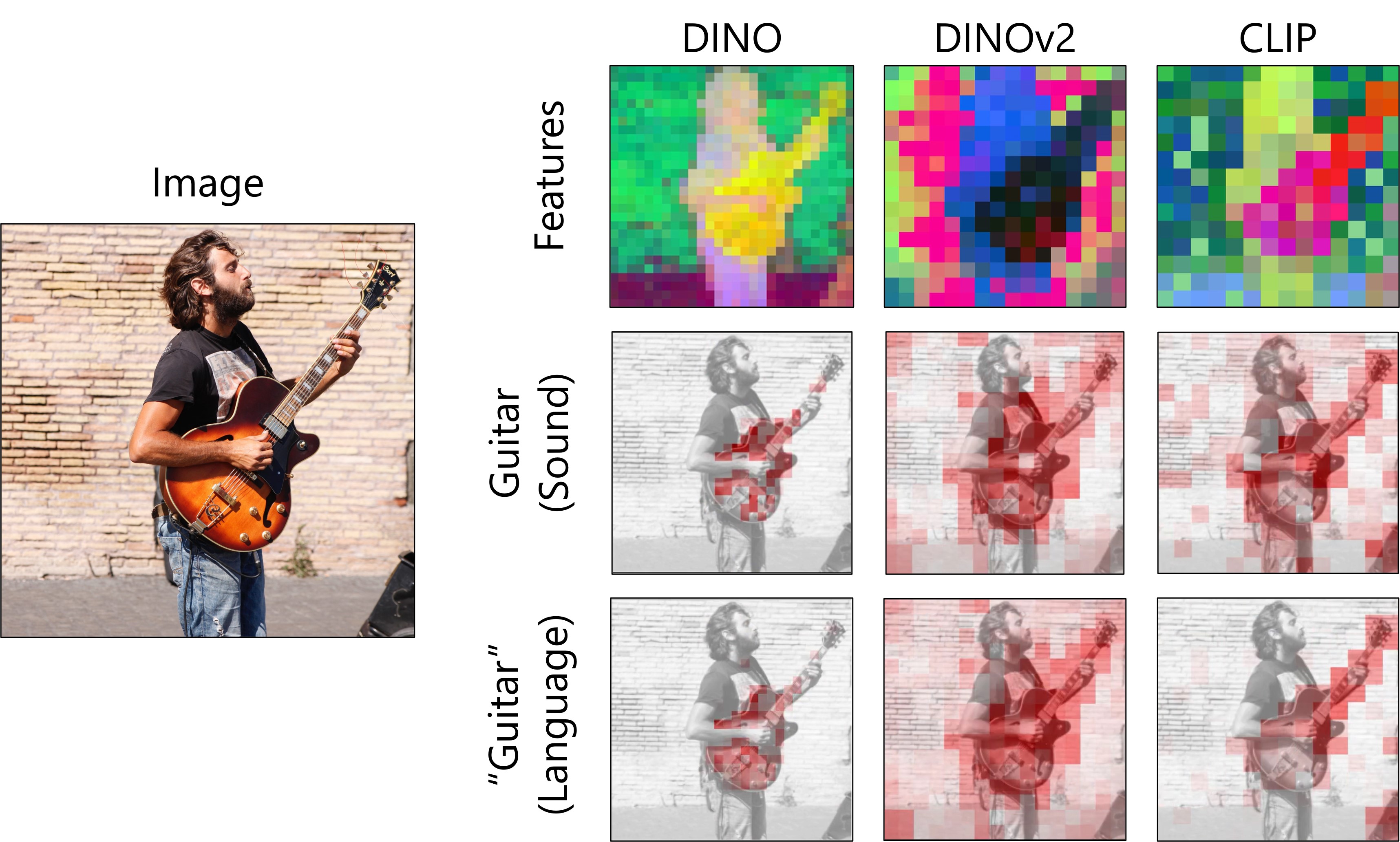}
    \caption{Comparison of sound and speech prompted localization of DenseAV with various choices of visual backbone. DINO's features are both the best for localization as well as the highest resolution because of its $8\times8$ patch size. }
    \label{fig:activation_comp_by_backbone}
\end{figure}

\newpage

\section{Associating Spoken Words to Visual Objects}

\begin{table}[h]
\centering
\begin{tabular}{c|ccccc}
\toprule
Visual Object & \multicolumn{5}{c}{Top 5 Retrieved Words}                    \\ \midrule
ottoman       & sofa         & chair    & chair     & seat       & living    \\
ruins         & brick        & stone    & castle    & clay       & stone     \\
dirt track    & dirt         & dirt     & trail     & field      & dirt      \\
monitor       & screen       & screen   & computer  & television & screen    \\
control panel & cockpit      & airplane & cockpit   & airplane   & airplane  \\
bar           & desk         & picture  & counter   & poker      & kitchen   \\
waterfall     & waterfall    & fountain & water     & waterfall  & waterfall \\
embankment    & trench       & land     & field     & land       & hill      \\
bleachers     & amphitheater & steps    & colosseum & step       & stairway  \\
snow          & snow         & snow     & snow      & mountain   & snow      \\ \bottomrule
\end{tabular}
\caption{ Top 5 word retrieval using DenseAV's visual object features on the speech prompted semantic segmentation dataset described in Section \ref{sec:speech-prompted}. We determine if DenseAV can perform fine-grained speech retrieval by seeing if inner activations properly highlight the definitions of objects. We average visual features of visual objects to form a visual object query vector. We then form word representations for the PlacesAudio validation set by averaging speech features over an utterance using word timing information provided by Microsoft's Speech to Text API. Feature averaging strategy is depicted in Figure~\ref{fig:averaging}. For each visual object, we retrieve the top 5 nouns from the PlacesAudio spoken captions. We do not average across words, so if a word appears twice in the table it represents two different spoken instances. Some visual objects are able to retrieve instances of speech that directly correspond to the name of the object, such as snow and waterfall. Others retrieve a variety of relevant words for example the ``ruins'' visual object retrieves instances of people saying ``brick'', ``castle'', and ``stone''. We note that the 10 visual objects selected were randomly selected from the hundreds in our speech prompted semantic segmentation dataset.  }
\label{tab:word_retrieval}

\end{table}

\begin{figure}[h]
    \centering
    \includegraphics[width=\linewidth]{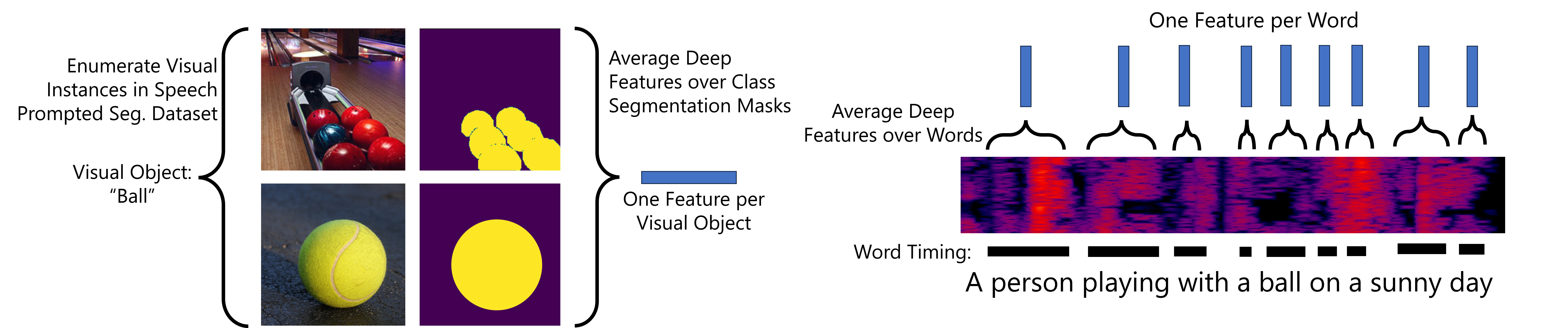}
    \caption{Diagram of feature averaging strategy used for the retrieval experiment in Table~\ref{tab:word_retrieval}. We average visual features over all instances of a visual object as shown in the left hand side, using the segmentation mask to only include visual features for the object of interest. To form features for each word in the places audio dataset, we use word timing information to average deep features over the extent of an utterance. Once we form features for all visual objects and all words, we retrieve the top 5 nouns for each visual object.}
    \label{fig:averaging}
\end{figure}

\newpage

\section{Failure Cases}


\begin{figure}[h]
    \centering
    \includegraphics[width=\linewidth]{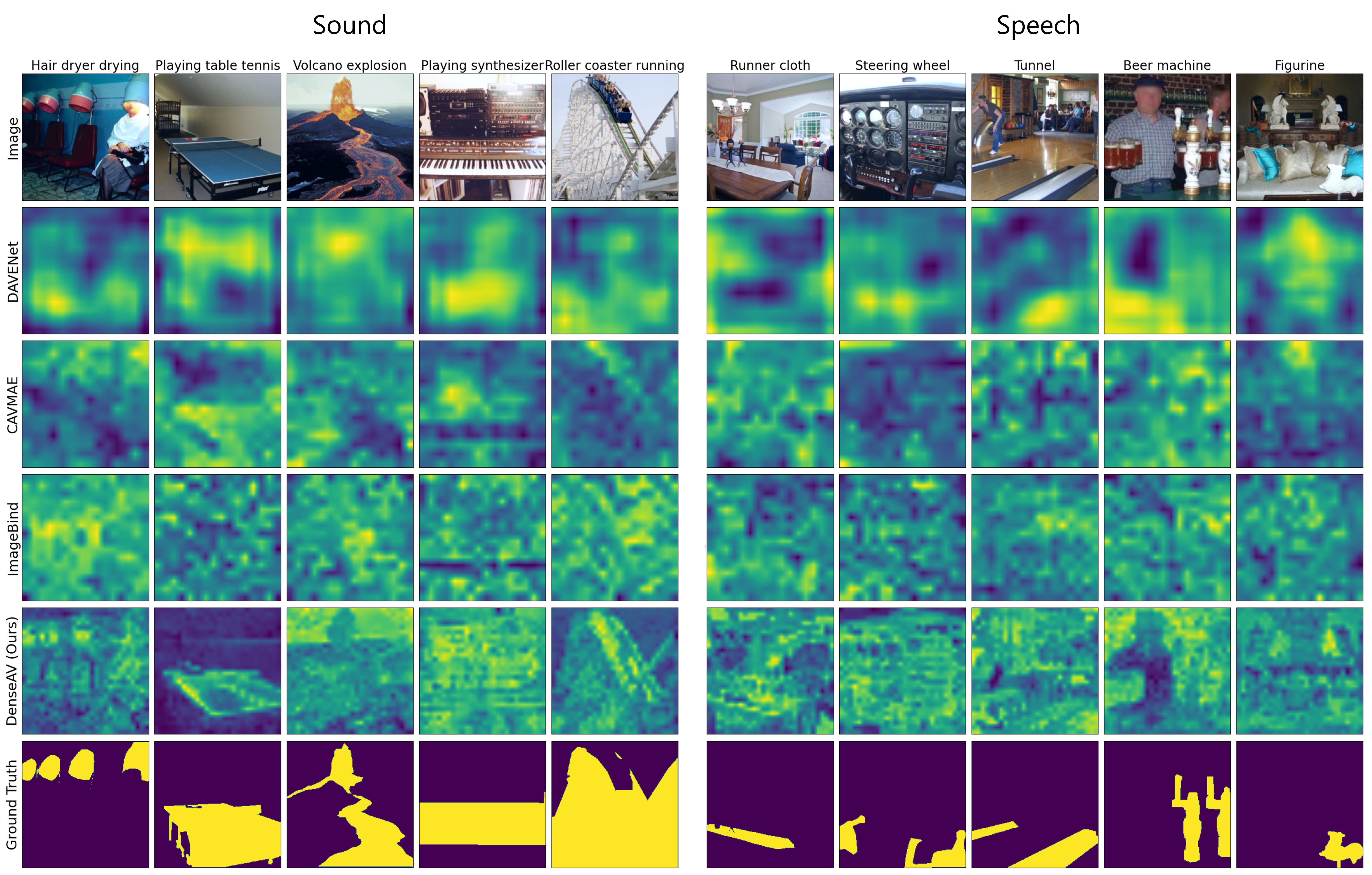}
    \caption{Examples of DenseAV's failure cases on speech and sound prompted semantic segmentation. On unusual visual objects such as the ``hair dryer drying'' activations are more diffuse than other hair dryers in the dataset, likely because of its rarer form. A similar effect appears in the steering wheel example likely because steering wheel is often infrequently used to describe airplane controls. Rare sounds like volcano explosions, or rare visual obnjects like the bowling ``tunnels'' cause similar diffuse activations.
    Like many discriminitive algorithms, DenseAV has some tendency to bias towards discriminitive regions such as the top of the table tennis board in the ``playing table tennis''. There is also some mismatch between ADE20K labels and what you might expect a reasonable algorithm should highlight, as evidenced by the ``roller coaster running'' sound example. Similarly in the ``Figurine'' example, the algorithm reasonably associates figurines with the lions in the background instead of the dog in the foreground. Finally the beer machine example shows how there's some ambiguity between whether an algorithm should respond to compound words and ideas. Should it couple ``beer'' to the beer glass and ``machine'' to  the spigots, or should ``beer-machine''  entirely couple to the spigots. DenseAV seems to choose the former as the beer in the foreground and background is also activated in this speech prompted example.  )}
    \label{fig:failure_cases}
\end{figure}

\newpage

\section{Comparing to DINO CLS Token Activations}

\begin{figure}[h]
    \centering
    \includegraphics[width=\linewidth]{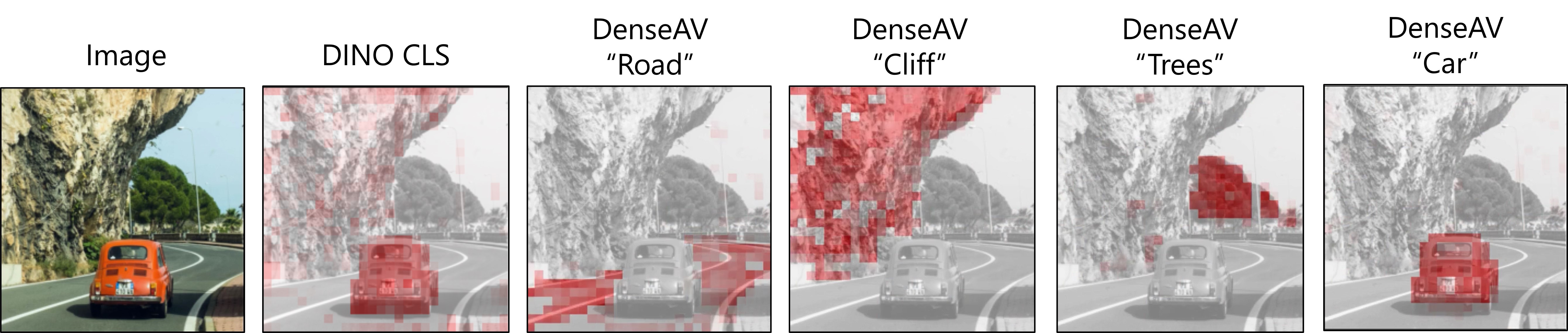}
    \caption{Comparison of DINO CLS token heatmap visualization~\cite{caron2021emerging} and DenseAV's activations. DenseAV does not just select salient objects as DINO's CLS token does. Instead, within a single video clip DenseAV can highlight the meaning of words as they are spoken. Depending on the word spoken, this can accurately highlight a variety of objects in the scene, even if they are less salient like the trees in the background. }
    \label{fig:dino_cls}
\end{figure}

\section{Visualizing Activations when an Object is not Present}

\begin{figure}[h]
    \centering
    \includegraphics[width=\linewidth]{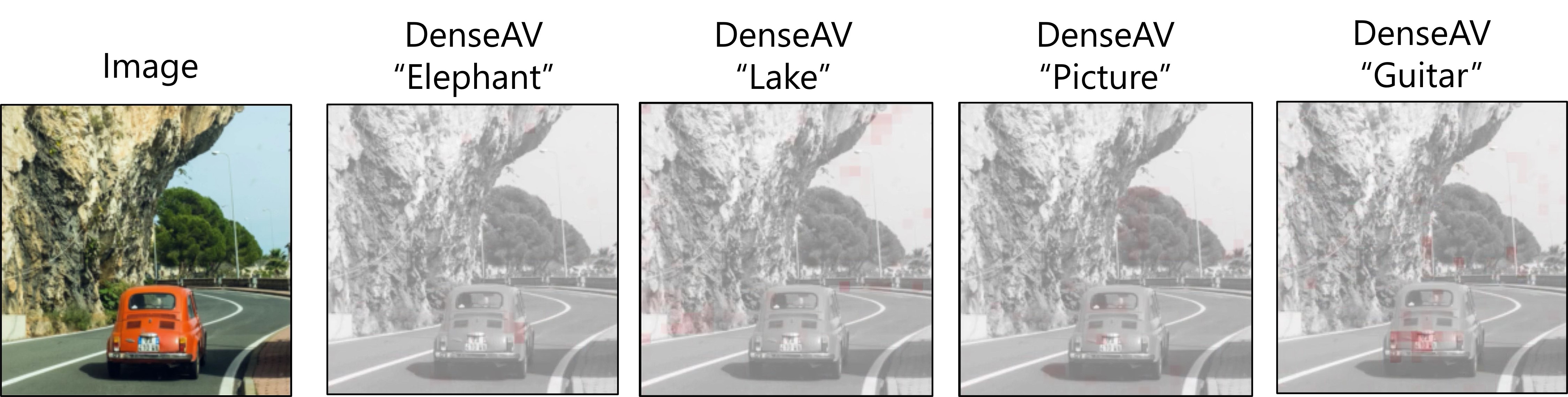}
    \caption{Visualization of DenseAV activations when an object is not present in a scene. DenseAV's activations are significantly smaller than when objects are present in a scene like in Figure~\ref{fig:dino_cls}.}
    \label{fig:negative_samples}
\end{figure}

\newpage

\section{Additional Regularizer Details}

\paragraph{Negative Audio Splicing} Though using \ref{eqn:contrastive} is enough to make a reasonable cross-modal retrieval system, the extreme flexibility of self-attention operator in modern transformers can lead to degenerate solutions. For example, we found that without regularizers that encourage local features to be meaningful, the network could develop its own ``global'' tokens by selecting a handful of local tokens to carry all of the information. This is similar to the observation of \cite{darcet2023vision} and we observed this occasionally in our audio branch, which would collapse to only use the first tokens. To keep the network from collapsing the semantics of the audio clip into a single token, we introduce small negative sample clips into our audio samples. These small negative audio regions are randomly spliced into the larger audio clip, and we encourage the network to set the couplings in these regions to zero with a $l_2$ regularizer. We include further details of the DenseAV's architecture, hyperparameters, and regularizers in the Supplement.

More formally, let $(a_b, v_b)_1^B$ be a \textbf{B}atch of $B$ paired audio and visual signals as before. Let $m_b \in [0,1]^T$ be a soft mask where that measures whether a given location in the audio signal is actually part of a spliced negative clip. For example, $m_b[t]=1$ when the clip at time $t$ is part of the negative clip, $m_b[t]=0$ in the positive part of the clip, and $0 < m_b[t] < 1$ in the small boundary regions when the true clip is being spliced into the negative clip and both sounds are present. Our negative audio splicing regularizer squares each entry of the similarity tensor and averages these according to the strength of the negative clip indicator $m_b$:

\begin{equation}
    \label{eqn:neg_splice}
    \mathcal{L}_{Splice} = \text{WeightedMean} (s(a_b, v_b)^2, m_b)
\end{equation}

Where the mean assumes that the weighting strength $m_b$ has been broadcast to the shape of $s(a_b, v_b)^2$. We point interested readers to the supplement for explicit formulations of these regularizers which are too verbose for the double-column format here. Intuitively, this term penalizes the network for having activations during a period of spliced negative audio. We also note that we apply this regularizer to any padded silence at the ends of short audio clips.

\paragraph{Calibration Regularization}

The calibration temperature provides the network with the crucial ability to increase or decrease its certainty by updating a single parameter. However, the network can also achieve this effect by increasing or decreasing the magnitudes of its features. We found that sometimes the temperature would accelerate downward, forcing the feature magnitudes to increase to compensate. As a result, the network would eventually saturate or become unstable. We hypothesize that this is due to optimizer momentum, and we prevent this ``runaway calibration'', by adding a small regularizer to the temperature parameter $\gamma$

\begin{equation}
	\mathcal{L}_{Cal} = \max(\log(1) - \log(\gamma), 0)^2
\end{equation} 

This term penalizes the calibrator when it drops below 1.0 and encourages the calibrator to stay at or above 1.0.

\paragraph{Nonnegative Pressure}

The InfoNCE loss function is invariant to the addition of a scalar to every inner product. Thus, to the network can choose to either find evidence of ``positive'' couplings connecting similar objects or ``negative'' couplings connecting regions that definitely do not belong together. We found that by encouraging the network to look for ``positive'' evidence, as opposed counterfactual evidence, improved training stability and performance across the key metrics we investigate. To encourage this behavior, we add a small regularizer to encourage inner products between features to be $\geq 0$. More specifically, let $\Omega$ be a set of 250 randomly selected coordinates $(b,b',k,f,t,h,w)$. We then form our non-negativity regularizer:

\begin{equation}
	\mathcal{L}_{NonNeg} = \frac{1}{|\Omega|} \sum_{\Omega}  \min \left(s(a_b, v_{b'})[k,f,t,h,w], 0 \right)^2
\end{equation} 

This regularizer penalizes the similarity tensor if it drops below zero, encouraging features to exhibit positive couplings. We note than other works \cite{hamilton2022unsupervised}, have noted the benefits of using only non-negative feature couplings.

\paragraph{Disentangement Regularization}

DenseAV's multi-head similarity aggregation allows the network to use its different heads to model different independent ways that the audio and video modalities could couple together. Interestingly we find that if we give DenseAV two heads, one naturally specializes to language and the other head to more generic sounds. In particular, we find that one head will rediscover the meaning of words by ``grounding'' them to visual objects and another head will localize which objects created a given sound. To purify this disentanglement of concepts without supervision, we encourage different attention heads of our algorithm to specialize. More specifically we penalize the network when multiple attention heads are simultaneously active. In our experiments we use two attention heads. As before, we let $(a_b, v_b)_1^B$ be a \textbf{B}atch of $B$ paired audio and visual signals. Our disentanglement loss for two heads is then:

\begin{equation}
	\mathcal{L}_{Dis} =\text{Mean}( |s(a_b, v_b)[1] \circ s(a_b, v_b)[2]|)
\end{equation} 

Where $\circ$ represents elementwise multiplication and $| \cdot |$ is the elementwise absolute value function. $[k]$ mirrors PyTorch slicing notation and refers to selecting the activations for only the $k$th attention head. Intuitively, this loss will encourage one head to be silent if the other head is active and can be viewed as a ``cross-term'' generalization of the $l^2$ regularizer \cite{hoerl1970ridge} for encouraging activation shrinkage.

\paragraph{Total Variation Smoothness}

To improve the quality and temporal consistency of discovered audio-visual couplings we impose a smoothness regularizer, $\mathcal{L}_{TV}$, in the audio-time dimension.

\begin{equation}
	\mathcal{L}_{TV} =\text{Mean}((\text{act}(1:t-1) - \text{act}(2:t))^2)
\end{equation} 

Where the activations for a given time slice $[1, t-1]$ are given by:

\begin{equation}
	\text{act}(1:t-1) = (s(a_b, v_b)[:,:,t',:,:])_{t'=1}^{t-1}
\end{equation} 

Informally, this regularizer penalizes when the inner product strengths change quickly over time.

\paragraph{Full Stability Regularizer}

Putting these terms together into a single equation we have:

\begin{equation}
    \mathcal{L}_{Stability} = \lambda_{Splice} \mathcal{L}_{Splice} + \lambda_{Cal} \mathcal{L}_{Cal} + \lambda_{NonNeg} \mathcal{L}_{NonNeg} + \lambda_{TV} \mathcal{L}_{TV}
\end{equation}

Where $ \lambda_{Splice}=0.01$,  $\lambda_{Cal}=0.1$, $\lambda_{NonNeg}=0.01$, and
$\lambda_{TV}=0.01$.

\newpage

\section{Regularizer Ablation}

\begin{table}[h]
\centering
\begin{tabular}{cccc|cc|cc}
\hline
\multicolumn{4}{c|}{Regularizer}                                                                                                                                            & \multicolumn{2}{c|}{Speech Semseg.}                 & \multicolumn{2}{c}{Places Acc. @ 10}                              \\
\multicolumn{1}{l}{$\mathcal{L}_{Cal}$} & \multicolumn{1}{l}{$\mathcal{L}_{NonNeg}$} & \multicolumn{1}{l}{$\mathcal{L}_{Splice}$} & \multicolumn{1}{l|}{$\mathcal{L}_{TV}$} & \multicolumn{1}{l}{mAP} & \multicolumn{1}{l|}{mIoU} & \multicolumn{1}{l}{$I   \to A$} & \multicolumn{1}{l}{$A   \to I$} \\ \hline
$\checkmark$                            & $\checkmark$                               & $\checkmark$                               & $\checkmark$                            & 48.7\%                  & 36.8\%                    & 94.2\%                          & 94.3\%                          \\
                                        & $\checkmark$                               & $\checkmark$                               & $\checkmark$                            & 49.1\%                  & 37.3\%                    & 94.3\%                          & 94.1\%                          \\
$\checkmark$                            &                                            & $\checkmark$                               & $\checkmark$                            & 48.2\%                  & 36.8\%                    & 94.1\%                          & 93.4\%                          \\
$\checkmark$                            & $\checkmark$                               &                                            & $\checkmark$                            & 48.6\%                  & 36.7\%                    & 94.8\%                          & 94.5\%                          \\
$\checkmark$                            & $\checkmark$                               & $\checkmark$                               &                                         & 49.0\%                  & 36.9\%                    & 94.2\%                          & 93.7\%                          \\
                                        &                                            &                                            &                                         & -                       & -                         & -                               & -                               \\ \hline
\end{tabular}
\caption{Ablation study of the different components of $\mathcal{L}_{Stability}$. We find that on the whole $\mathcal{L}_{Stability}$ is needed to avoid collapse as shown the the bottom row of the table. However, removing any individual term does not have much effect on the final metrics.}
\end{table}


\end{document}